\documentclass[pdflatex,sn-basic,Numbered]{sn-jnl}

\usepackage{graphicx}
\usepackage{amsmath,amssymb,amsfonts}
\usepackage{booktabs}
\usepackage{xcolor}
\usepackage{textcomp}
\usepackage{url}
\makeatletter
\def\@seccntformat#1{}
\makeatother

\begin{document}

\title[Opportunistic Bone-Loss Screening from Knee Radiographs]{Opportunistic Bone-Loss Screening from Routine Knee Radiographs Using a Multi-Task Deep Learning Framework with Sensitivity-Constrained Threshold Optimization}


\author[1]{\fnm{Zhaochen} \sur{Li}}\equalcont{These authors contributed equally as co-first authors.}
\author[1]{\fnm{Xinghao} \sur{Yan}}\equalcont{These authors contributed equally as co-first authors.}
\author[1]{\fnm{Runni} \sur{Zhou}}
\author[1]{\fnm{Xiaoyang} \sur{Li}}
\author[1]{\fnm{Chenjie} \sur{Zhu}}
\author[1]{\fnm{Gege} \sur{Wang}}
\author[2]{\fnm{Yu} \sur{Shi}}
\author[3]{\fnm{Lixin} \sur{Zhang}}
\author[1]{\fnm{Rongrong} \sur{Fu}}
\author[1]{\fnm{Liehao} \sur{Yan}}
\author*[1,4]{\fnm{Yuan} \sur{Chai}}
\email{chaiyuan@neu.edu.cn}

\affil[1]{\orgname{ORBIT Lab, College of Medicine and Biological Information Engineering, Northeastern University}, \orgaddress{\city{Shenyang}, \state{Liaoning}, \country{China}}}

\affil[2]{\orgname{Department of Radiology, Liaoning Provincial Key Laboratory of Medical Imaging, Liaoning Provincial Key Laboratory of Imaging Technology and Artificial Intelligence, Shengjing Hospital of China Medical University}, \orgaddress{\city{Shenyang}, \state{Liaoning}, \country{China}}}

\affil[3]{\orgname{Rehabilitation Center, Liaoning Provincial Key Laboratory of Medical Imaging, Liaoning Provincial Key Laboratory of Imaging Technology and Artificial Intelligence, Shengjing Hospital of China Medical University}, \orgaddress{\city{Shenyang}, \state{Liaoning}, \country{China}}}

\affil[4]{\orgname{The University of Sydney, Sydney Musculoskeletal Health and the Kolling Institute, Northern Clinical School, Faculty of Medicine and Health and the Northern Sydney Local Health District}, \orgaddress{\city{Sydney}, \state{NSW}, \country{Australia}}}


\abstract{\textbf{Background:}
Osteoporosis and osteopenia are often undetected until fracture. Knee radiographs are routinely acquired and may support opportunistic screening.

\textbf{Objective:}
To develop and evaluate STR-Net, a multi-task deep learning framework for opportunistic bone-loss screening from knee radiographs.

\textbf{Methods:}
STR-Net was developed on 1,570 radiographs from two publicly redistributed datasets and trained for binary screening, osteopenia/osteoporosis sub-classification, and exploratory T-score regression. A sensitivity-constrained threshold selected during development was fixed ($\tau^{\star}=0.44$) and transferred without re-tuning to an independent external cohort with  DXA-based hip or spine BMD and corresponding derived T-scores.

\textbf{Results:}
On the held-out test set (n=224), STR-Net achieved AUROC 0.933 and AUPRC 0.956 for binary screening, with sensitivity 0.904 and specificity 0.773 at $\tau^{\star}=0.44$. Severity sub-classification AUROC was 0.898. The pilot T-score branch (n=31) showed Pearson $r=0.801$ (MAE 0.279; RMSE 0.347). In the external cohort (n=42), screening AUROC was 0.878 (95\,\% CI 0.719--0.983), with sensitivity 0.800 and specificity 0.750; T-score correlation was $r=0.766$. Development labels were heterogeneous and not uniformly DXA-based; reference-standard-anchored evaluation was therefore restricted to the external cohort.

\textbf{Conclusions:}
STR-Net provides preliminary evidence for opportunistic bone-loss screening from knee radiographs under heterogeneous development-label conditions. Prospective validation and a DXA-matched development cohort are required before deployment.}


\maketitle

\section{Introduction}\label{sec:intro}

Osteoporosis and osteopenia are common, clinically important, and frequently underdiagnosed until fracture~\cite{who1994,kanis2019,hernlund2013,siris2004}. Although DXA is the reference standard, broad screening remains limited by access and workflow constraints~\cite{shepstone2018}. Opportunistic screening therefore aims to extract bone-health signals from examinations already acquired for other indications~\cite{pickhardt2013,yamamoto2020}.

Most opportunistic osteoporosis work has focused on CT, while plain radiography is less explored despite far wider availability~\cite{pickhardt2013,pan2020,fang2021,yamamoto2020,zhang2022xray,hsieh2021nature}. Knee radiographs are especially attractive because they are high-volume in older adults and contain periarticular cancellous bone that can reflect systemic bone-loss patterns~\cite{bedson2007kneexray,riggs1981jci}. Prior knee-radiograph deep learning studies have mainly addressed osteoarthritis and related tasks~\cite{tiulpin2018kneeoa}; screening for systemic bone loss from the same images is comparatively underreported.

Existing radiograph-based screening pipelines are often single-task and use a default 0.5 threshold, which is not optimal for screening where missing positive cases is costly. We therefore developed STR-Net, a shared-backbone multi-task model that outputs: (i) binary screening (Normal vs.\ Bone Loss), (ii) severity sub-classification (Osteopenia vs.\ Osteoporosis within Bone Loss), and (iii) exploratory T-score regression. The primary decision threshold is selected with an explicit sensitivity constraint to align with referral-oriented clinical use.

The system addresses the following design considerations:

\begin{enumerate}
  \item \textbf{Binary screening with sensitivity constraint.} The primary task classifies Normal vs.\ Bone Loss (Osteopenia $\cup$ Osteoporosis), with threshold selection enforcing sensitivity $\geq 0.86$ while maximizing specificity.

  \item \textbf{Auxiliary severity and T-score outputs.} A masked severity branch estimates Osteopenia vs.\ Osteoporosis, and a weakly coupled branch provides exploratory T-score regression with optional age/sex/BMI fusion.

  \item \textbf{Task-aware representation routing (TARR).} Task-specific embeddings are derived from shared features to reduce negative interference between screening and auxiliary heads.
\end{enumerate}

The key contributions are a knee-radiograph multi-task screening framework, a sensitivity-constrained threshold policy for referral settings, and a unified ablation/benchmark analysis under a fixed split protocol. Sections~\ref{sec:methods}--\ref{sec:conclusions} report methods, results, interpretation, and limitations.

\section{Materials and Methods}\label{sec:methods}

\subsection{Data Sources}\label{subsec:dataset}

\begin{figure}[h]
\centering
\includegraphics[width=1.0\linewidth]{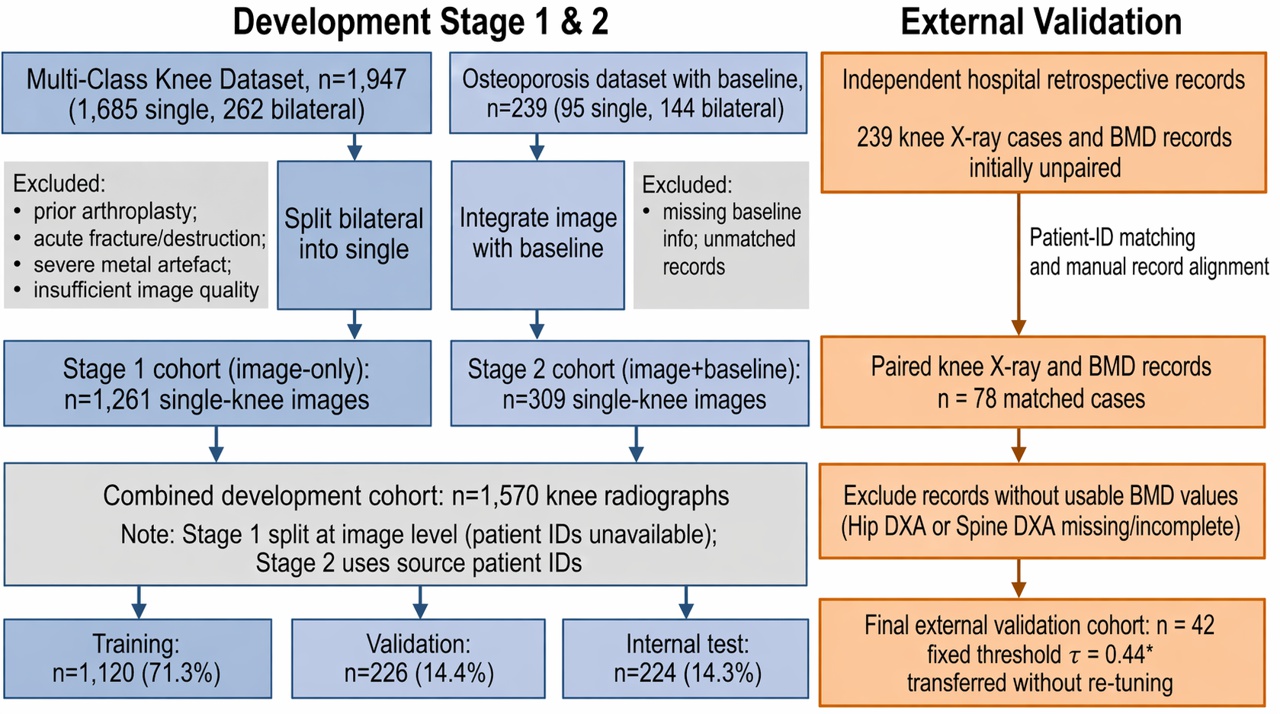}

\caption{Development and external-validation workflow as rendered in the final figure. Stage~1 starts from the Multi-Class Knee dataset (n=1{,}947; 1{,}685 single-knee and 262 bilateral images), applies exclusion criteria, and splits bilateral studies into single-knee images to form the Stage~1 cohort (n=1{,}261). Stage~2 starts from the osteoporosis dataset with baseline data (n=239; 95 single and 144 bilateral), integrates images with baseline metadata after exclusions to form the Stage~2 cohort (n=309). The combined development cohort (n=1{,}570) is split into training (n=1{,}120), validation (n=226), and internal test (n=224). The independent external cohort is finalized at n=42 and evaluated at fixed $\tau^{\star}=0.44$ without re-tuning.}\label{fig:patient_flow}
\end{figure}

We used two fully de-identified public knee-radiograph repositories redistributed on Kaggle: Stage~1 (``Osteoporosis X-ray'', linked to Wani and Arora~\cite{wani2022qus}) and Stage~2 (``Multi-class Knee Osteoporosis X-ray Dataset''~\cite{sarhan2024multiclass}). Image-level inclusion required a weight-bearing antero-posterior knee radiograph with an unambiguous source-repository class label (Normal, Osteopenia, or Osteoporosis). Exclusion criteria were prior arthroplasty/prosthesis, acute fracture or destructive change, severe metal artefact, and inadequate image quality. Figure~\ref{fig:patient_flow} summarizes the full workflow shown in the final figure, including source-pool counts, bilateral-to-single processing, development splitting, and the fixed-threshold external-validation branch.

A total of 1,570 development knee radiographs were included after data cleaning. The dataset was assembled from two stages:

\begin{itemize}
  \item \textbf{Stage~1} (n=1,261): labels are repository-assigned at directory level; the linked source reports calcaneal QUS-derived T-scores rather than DXA~\cite{wani2022qus}.
  \item \textbf{Stage~2} (n=309): explicit class labels (\texttt{label\_cls}: 0/1/2), continuous T-scores, and baseline covariates (age, sex, BMI)~\cite{sarhan2024multiclass}; label provenance remains heterogeneous because Stage-1-derived material is included.
\end{itemize}

The two stages were merged on image path after case-insensitive deduplication. The distribution of diagnostic categories in the combined dataset is shown in Table~\ref{tab:data_distribution}.

\begin{table}[h]
\normalsize
\begin{tabular*}{\linewidth}{@{}l@{\extracolsep{\fill}}cccc@{}}
\toprule
\textbf{Category} & \textbf{Train} & \textbf{Val} & \textbf{Test} & \textbf{Total} \\
\midrule
Normal      & 279  & 60  & 88 & 427 \\
Bone Loss   & 841  & 166 & 136 & 1,143 \\
Osteopenia   & 450 & 79 & 65 & 594 \\
Osteoporosis & 391 & 87 & 71 & 549 \\
\midrule
\textbf{Total}  & 1,120 & 226 & 224 & 1,570 \\
\bottomrule
\end{tabular*}
\caption{Dataset composition by diagnostic category and data split. Bone Loss denotes the union of Osteopenia and Osteoporosis. Counts were recomputed from the merged Stage~1 and Stage~2 split files after case-insensitive path deduplication.}\label{tab:data_distribution}
\end{table}

Development labels were repository-assigned and heterogeneous (Stage~1 QUS-linked directory labels; Stage~2 redistribution with mixed provenance). No internal DXA re-labeling was performed. WHO boundaries were retained for category mapping~\cite{who1994,kanis2019}.

Stage~1 lacks verifiable patient identifiers, so Stage~1 splitting was image-level; residual identity leakage cannot be excluded if multiple radiographs from one patient exist. Stage~2 split files include patient identifiers and were retained as provided.

Perceptual-hash auditing (64-bit pHash; Hamming distance $\leq 4$) was performed across splits as a data-quality check.

Demographic characteristics are reported in Section~\ref{subsec:baseline}.

To test generalization beyond public development data, we assembled an independent retrospective cohort with paired weight-bearing AP knee radiography and DXA-based hip or spine BMD within a predefined clinical window at a collaborating hospital. Bone status followed WHO thresholds~\cite{who1994,kanis2019}: Normal (T-score $\geq -1.0$), Osteopenia ($-2.5 <$ T-score $< -1.0$), and Osteoporosis (T-score $\leq -2.5$); for screening, Osteopenia and Osteoporosis were merged as Bone Loss. Ethical approval was obtained from the institutional Research Ethics Committee (Approval No.: 2025PS1425K), and the study followed the Declaration of Helsinki.

External images used the same preprocessing pipeline (Section~\ref{subsec:preprocess}). Model weights were frozen, no external fine-tuning was performed, and the internal validation threshold $\tau^{\star}$ from~\eqref{eq:threshold_opt} was transferred without re-estimation. Screening probabilities and auxiliary T-score outputs were compared with DXA-derived binary labels and continuous hip- or spine-derived T-scores. Reference standards are asymmetric: development labels are repository-assigned and partly QUS-linked, whereas external labels are DXA-based; therefore, external results are interpreted as transfer evidence for risk-related imaging signal rather than calibration-grade site-specific DXA prediction~\cite{riggs1981jci}.
\subsection{Image preprocessing and augmentation}\label{subsec:preprocess}

All images were converted to 8-bit grayscale. The preprocessing pipeline consisted of the following steps:

\begin{enumerate}
  \item \textbf{Foreground cropping.} An Otsu-based thresholding method was applied to detect the foreground region, followed by cropping with a 3\% margin to remove extraneous background and collimation artifacts. A safe fallback to the original image dimensions was employed when the detected foreground was smaller than 10\% of the original image area.

  \item \textbf{Intensity normalization.} Pixel intensities were clipped to the 1st--99th percentile range and rescaled to 0--255 to harmonize intensity distributions across different acquisition protocols.

  \item \textbf{Resizing.} All images were resized to $512 \times 512$ pixels using bilinear interpolation.

  \item \textbf{Tensor normalization.} Pixel values were normalized using a mean of 0.5 and standard deviation of 0.25 for the single grayscale channel.
\end{enumerate}

During training, the following augmentation transforms were applied:

\begin{itemize}
  \item Random horizontal flip (probability = 0.5).
  \item Random affine transformation (rotation $\leq$ 8$^\circ$, translation $\leq$ 4\%, scale 0.95--1.05).
  \item Random border masking (probability = 0.5): Random strips of up to 4\% of image dimensions along the borders were replaced with the median pixel value, reducing the model's reliance on edge artifacts.
\end{itemize}

No augmentation was applied during validation or testing.

\subsection{Model Architecture}\label{subsec:model}

\begin{figure}[t]
\centering
\includegraphics[width=\linewidth]{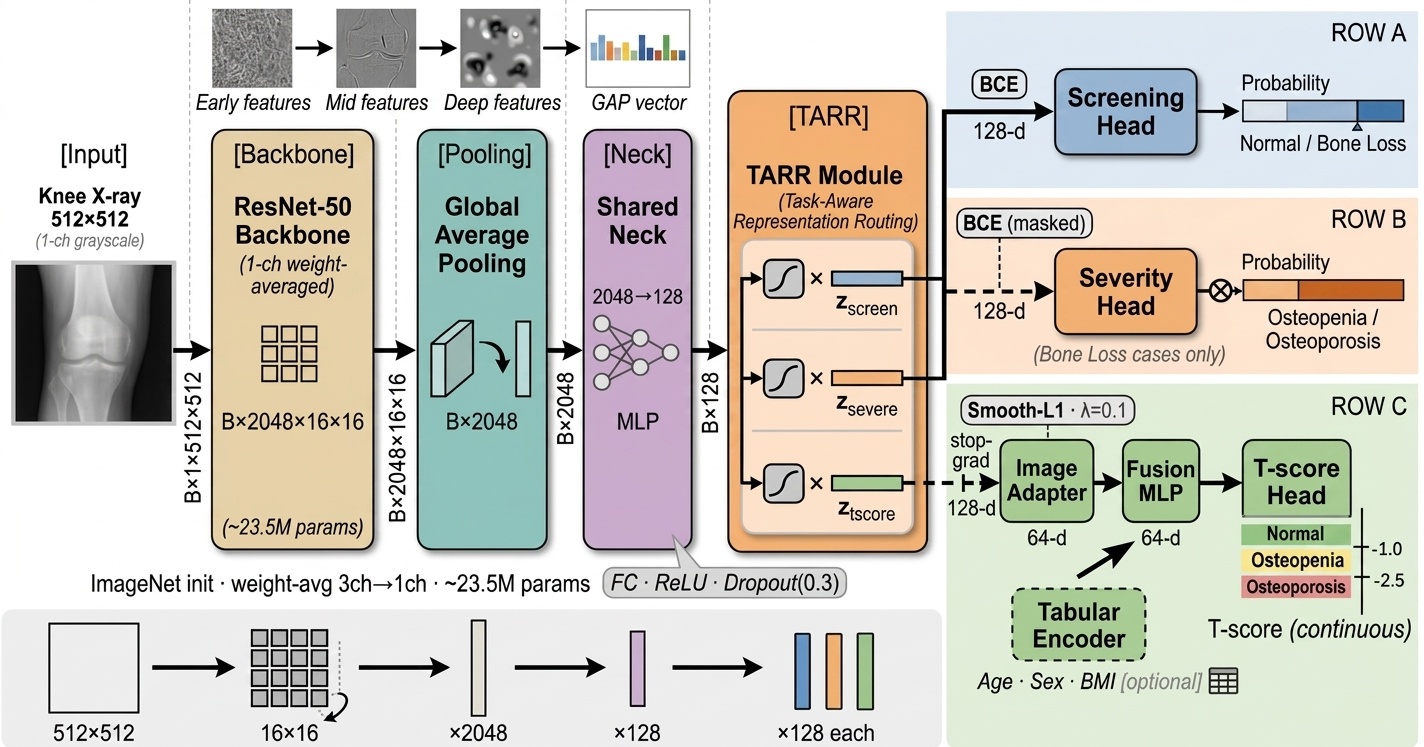}
\caption{STR-Net architecture for opportunistic bone-loss screening from routine knee radiographs. A ResNet-50 backbone extracts image features, which are routed by the task-aware representation routing (TARR) module into three parallel branches: bone-loss screening (Normal vs.\ Bone Loss), severity sub-classification (Osteopenia vs.\ Osteoporosis, applied only to bone-loss cases), and T-score regression with optional clinical feature fusion (pilot evaluation; Section~\ref{subsec:tscore}). The three-output design allows simultaneous screening and grading from a single radiograph without additional imaging or DXA referral.}\label{fig:architecture}
\end{figure}

\begin{figure}[t]
\centering
\includegraphics[width=\linewidth]{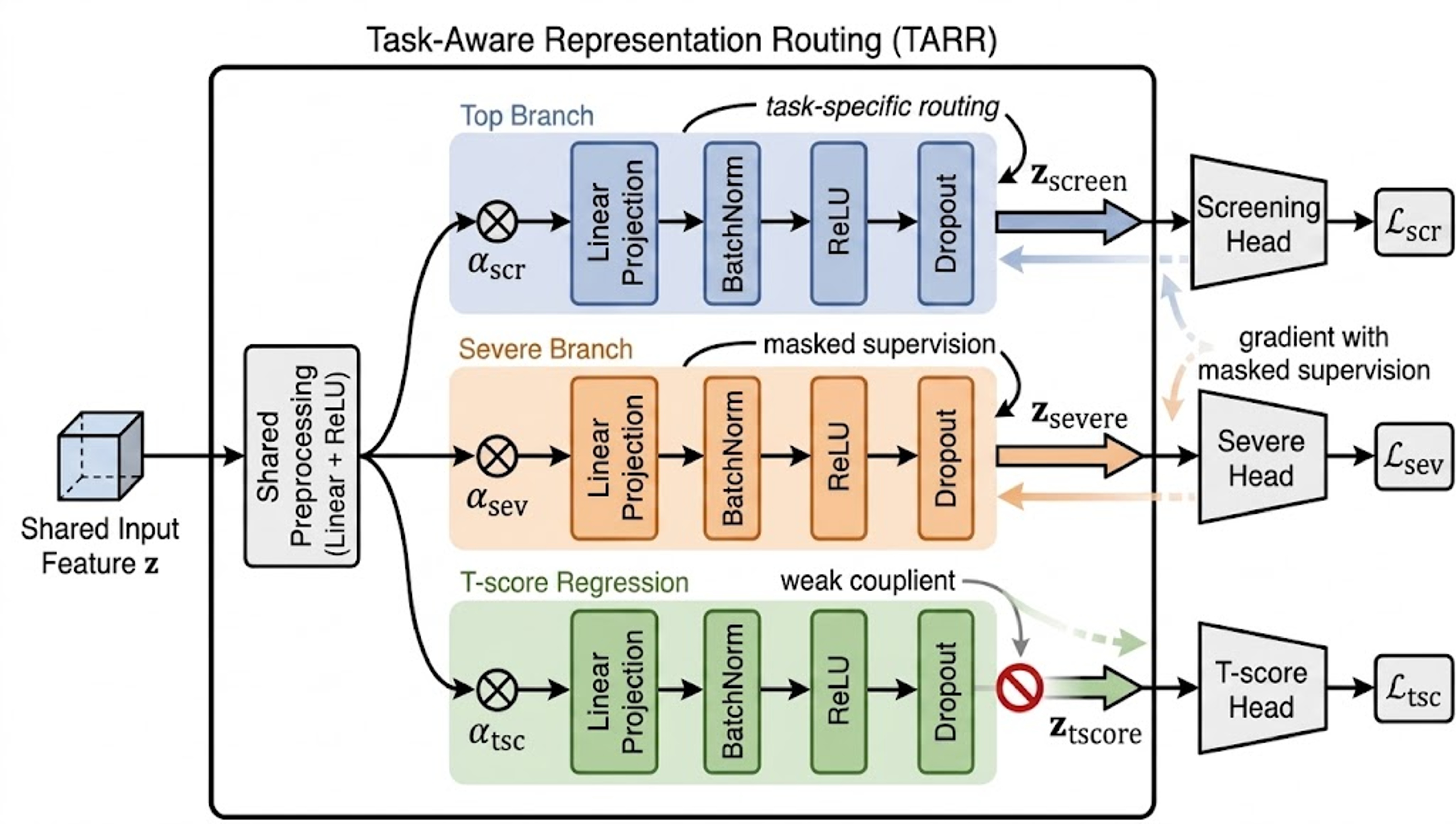}
\caption{Task-aware representation routing (TARR) module. A shared image feature is split into three task-specific branches: bone-loss screening, severity sub-classification (restricted to bone-loss cases), and T-score regression (weakly coupled to avoid interference with screening). Each branch generates an independent prediction, allowing the model to simultaneously screen for bone loss, grade its severity, and output an auxiliary T-score estimate from a single radiograph (evaluated only on the small labeled subset; Section~\ref{subsec:tscore}).}\label{fig:tarr_detail}
\end{figure}

Figure~\ref{fig:architecture} summarizes the shared-backbone pipeline, and Fig.~\ref{fig:tarr_detail} details TARR routing. A ResNet-50~\cite{he2016resnet} backbone (adapted for single-channel input) produces global image features, followed by global average pooling and a shared neck projection with ReLU and dropout~\cite{srivastava2014dropout}. In the final TARR diagram, each branch applies task-wise scaling and a per-branch Linear$\rightarrow$BatchNorm$\rightarrow$ReLU$\rightarrow$Dropout projection to produce task embeddings for screening, severity, and T-score outputs~\cite{caruana1997multitask,ruder2017overview}. Screening uses BCE, severity uses masked BCE on Bone-Loss cases only, and the T-score path is weakly coupled via stop-gradient before image-adapter and fusion-MLP regression.

\subsection{Training Strategy}\label{subsec:training}

Training ran up to 100 epochs with early stopping (patience 30, monitored by validation AUPRC). The total loss was

\begin{equation}
\mathcal{L} = \mathcal{L}_{\text{screen}} + \lambda_s \cdot \mathcal{L}_{\text{severe}} + \lambda_t \cdot \mathcal{L}_{\text{tscore}}
\label{eq:loss}
\end{equation}

\noindent where $\lambda_s=0.3$ and $\lambda_t=0.1$. $\mathcal{L}_{\text{screen}}$ and $\mathcal{L}_{\text{severe}}$ were weighted BCE objectives (severity masked to Bone-Loss samples), and $\mathcal{L}_{\text{tscore}}$ was Smooth-L1 loss over samples with available T-scores.
Focal loss~\cite{lin2017focal} was not used because class imbalance was moderate. The T-score loss was activated after a 5-epoch warmup to avoid destabilizing early screening optimization.

Optimization used AdamW~\cite{loshchilov2019adamw} (learning rate $1\times10^{-4}$, weight decay $1\times10^{-4}$), ReduceLROnPlateau (factor 0.5, patience 20), and gradient clipping (max norm 1.0). Seed was fixed at 42. Training was run on one NVIDIA GPU (typically under 6 hours/run). Total trainable parameters were $\approx$23.95\,M. Hyperparameters are listed in Table~\ref{tab:hyperparams}.

\begin{table}[h]
\normalsize
\caption{Key training hyperparameters and model configuration.}\label{tab:hyperparams}
\begin{tabular*}{\linewidth}{@{}l@{\extracolsep{\fill}}l@{}}
\toprule
\textbf{Hyperparameter} & \textbf{Value} \\
\midrule
Backbone                      & ResNet-50 \\
Input resolution              & 512 $\times$ 512 pixels, 1 channel \\
Backbone pre-training         & ImageNet~\cite{russakovsky2015imagenet} \\
First-conv channel adaptation & Weight averaging (3-ch $\to$ 1-ch) \\
Neck: FC output dim           & 128 \\
Neck: Dropout rate            & 0.3 \\
TARR: adapter hidden dim      & 64 \\
TARR: branch output dim       & 128 \\
T-score image adapter dim     & 128 $\to$ 64 \\
T-score tabular encoder dim   & 3 $\to$ 32 $\to$ 32 \\
T-score fusion MLP dim        & 96 $\to$ 64 $\to$ 1 \\
Optimizer                     & AdamW \\
Learning rate                 & $1 \times 10^{-4}$ \\
Weight decay                  & $1 \times 10^{-4}$ \\
LR scheduler                  & ReduceLROnPlateau (factor=0.5, patience=20) \\
Gradient clipping (max norm)  & 1.0 \\
Max epochs                    & 100 \\
Early stopping patience       & 30 epochs (monitored: val AUPRC) \\
Batch size                    & 32 \\
$\lambda_s$ (severe weight)   & 0.3 \\
$\lambda_t$ (T-score weight)  & 0.1 \\
T-score warmup                & 5 epochs \\
Random seed                   & 42 \\
Total trainable parameters    & $\approx$23.95\,M \\
\bottomrule
\end{tabular*}
\end{table}

\subsection{Outcomes, evaluation metrics, threshold selection, and Grad-CAM}\label{subsec:evaluation}

The primary endpoint was binary screening performance (Normal vs.\ Bone Loss) on the held-out internal test set at a sensitivity-constrained operating threshold. Secondary endpoints were severity sub-classification within Bone Loss, exploratory T-score regression, and qualitative Grad-CAM localization.

Primary discrimination metric was AUPRC, with AUROC, accuracy, balanced accuracy, sensitivity, and specificity reported as complements~\cite{saito2015precision}. Severity used the same classification metrics within the Bone-Loss subset. T-score performance was summarized by MAE, RMSE, and Pearson correlation.

Reporting followed TRIPOD+AI~\cite{collins2024tripod}. Continuous variables are shown as mean~$\pm$~SD and compared with Kruskal--Wallis tests; categorical variables are shown as counts/percentages and compared with chi-square tests. Screening CIs (internal and external cohorts) used stratified bootstrap resampling (2{,}000 replicates)~\cite{efron1994bootstrap}. Pearson-$r$ CIs used Fisher-$z$ transformation. Two-sided $p<0.05$ was considered statistically significant.

To align with screening use, threshold selection prioritized high sensitivity. During development, we used a sensitivity-constrained grid search on the validation set. Let $\mathrm{Sens}(\tau)$ and $\mathrm{Spec}(\tau)$ denote sensitivity and specificity at threshold $\tau$ on $\hat{p}_{\text{screen}}$, with grid $\mathcal{G}=\{0.20,0.21,\ldots,0.80\}$:
\begin{equation}
\tau^{\star} \in \arg\max_{\tau \in \mathcal{G}_{\mathrm{feas}}} \mathrm{Spec}(\tau),\qquad
\mathcal{G}_{\mathrm{feas}} = \bigl\{\tau \in \mathcal{G} : \mathrm{Sens}(\tau) \geq 0.86\bigr\}.
\label{eq:threshold_opt}
\end{equation}
The selected threshold was $\tau^{\star}=0.44$, transferred without re-tuning to the internal test and external cohorts. Figure~\ref{fig:roc_pr}(c) reports internal test operating-point metrics at this fixed threshold (sensitivity 0.904, specificity 0.773).

Grad-CAM~\cite{selvaraju2017gradcam} was computed from the last ResNet-50 convolutional block for representative test cases, upsampled to input resolution, and overlaid on radiographs with a jet colormap. Anatomical interpretation was qualitative and focused on periarticular cancellous bone in the distal femur and proximal tibia~\cite{riggs1981jci}. Implementation used Python~3.10, PyTorch~2.1~\cite{paszke2019pytorch}, and standard scientific libraries.

\section{Results}\label{sec:results}

\subsection{Baseline Characteristics}\label{subsec:baseline}

Table~\ref{tab:demographics} reports Stage~2 baseline characteristics (n=309), the only subset with linked age/sex/BMI metadata. Stage~1 (n=1,261) lacks individual demographic records, so a full-cohort demographic table is not available. In Stage~2, osteoporosis cases were older than osteopenia and normal groups (62.7 vs.\ 51.5 vs.\ 34.5 years; $p<0.001$), female proportion differed by category ($p=0.001$), and BMI difference was marginal ($p=0.051$). Repository-provided T-scores also differed across categories ($p<0.001$), supporting internal label consistency within Stage~2.

\begin{table}[h]
\normalsize
\begin{tabular*}{\linewidth}{@{}l@{\extracolsep{\fill}}cccc@{}}
\toprule
\textbf{Characteristic} & \textbf{Normal} & \textbf{Osteopenia} & \textbf{Osteoporosis} & \textbf{$p$-value} \\
 & (n=34) & (n=206) & (n=69) & \\
\midrule
Age (years)       & $34.5 \pm 7.0$ & $51.5 \pm 8.9$ & $62.7 \pm 12.0$ & $<$0.001 \\
Female sex, $n$~(\%) & 20 (58.8\%) & 130 (63.1\%) & 26 (37.7\%) & 0.001 \\
BMI (kg/m$^2$)    & $26.69 \pm 3.77$ & $28.07 \pm 4.09$ & $26.95 \pm 3.76$ & 0.051 \\
T-score (source repository) & $-0.66 \pm 0.21$ & $-1.88 \pm 0.38$ & $-2.68 \pm 0.16$ & $<$0.001 \\
\bottomrule
\end{tabular*}
\caption{Baseline characteristics of the Stage~2 subgroup (n=309) by diagnostic category. Continuous variables: mean~$\pm$~SD (Kruskal--Wallis); categorical variables: $n$~(\%) (chi-square).}\label{tab:demographics}
\end{table}

\subsection{Training Convergence}\label{subsec:convergence}

\begin{figure}[t]
\centering
\includegraphics[width=\linewidth]{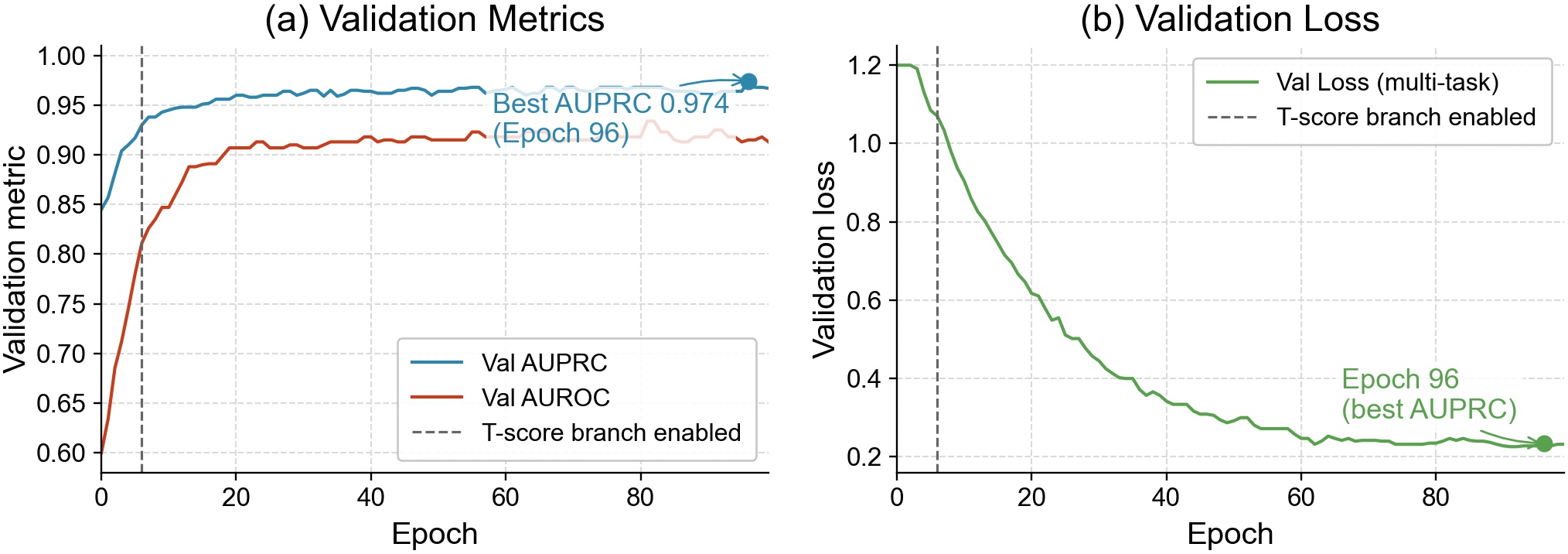}
\caption{Training convergence over 100 epochs. \textbf{(a)}~Validation AUPRC (blue) and AUROC (red) with the T-score-branch activation point marked by the dashed vertical line; best AUPRC = 0.974 at epoch 96. \textbf{(b)}~Multi-task validation loss decreases monotonically without instability at branch activation.} \label{fig:training_curves}
\end{figure}

The model was trained for 100 epochs without triggering early stopping. The best validation AUPRC of 0.974 was achieved at epoch 96 (val AUROC = 0.927; Fig.~\ref{fig:training_curves}). Enabling the T-score branch after a short warm-up (dashed vertical marker) produced no visible instability, and the learning rate remained constant throughout training, indicating that ReduceLROnPlateau was not triggered within the 100-epoch budget.

\subsection{Primary Screening Performance}\label{subsec:screening_perf}

\begin{figure}[h]
\centering
\includegraphics[width=\linewidth]{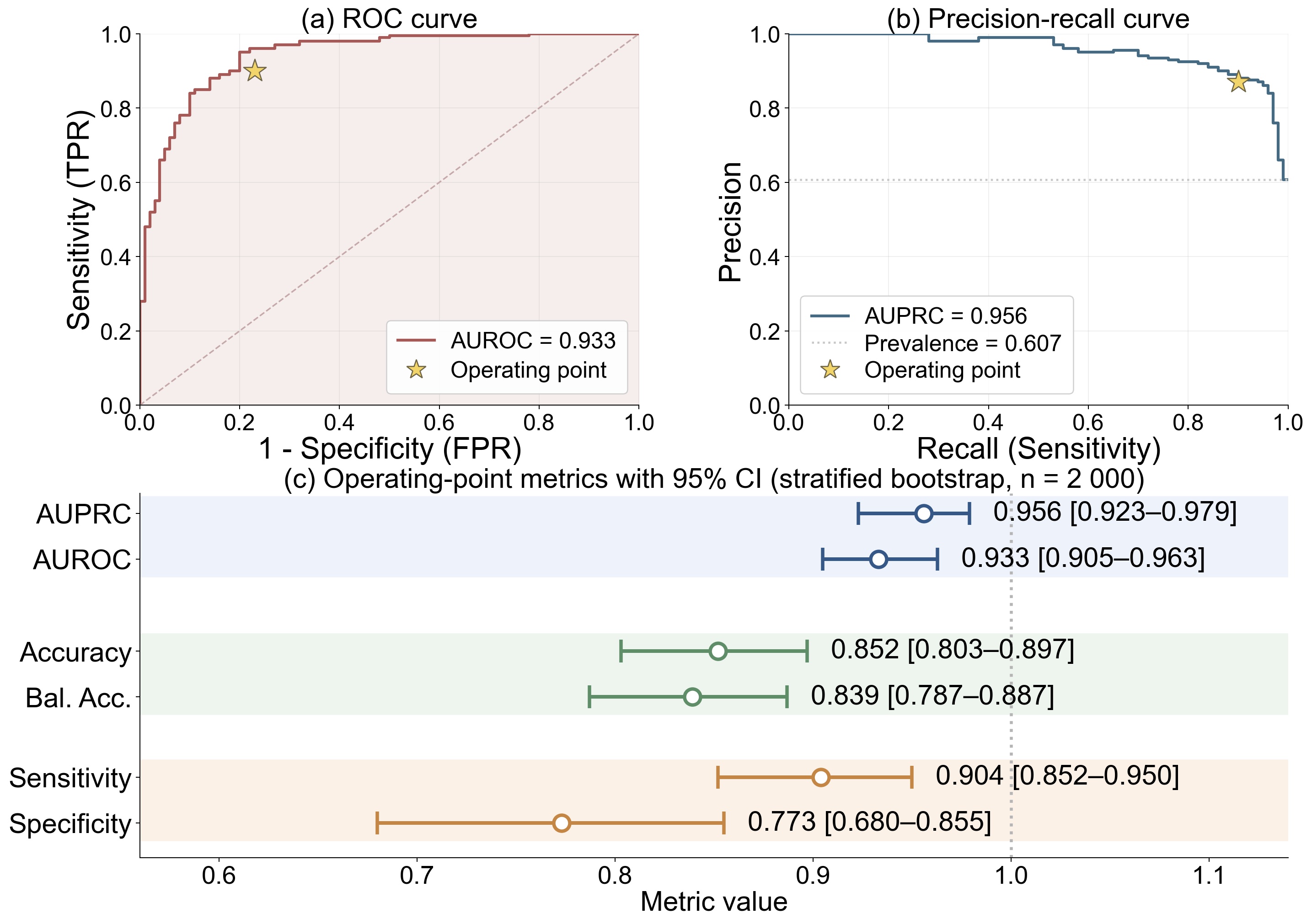}
\caption{ROC curve \textbf{(a)}, precision--recall curve \textbf{(b)}, and operating-point metrics with 95\% CI \textbf{(c)} for the primary screening head on the held-out test set (n=224; 2\,000 stratified bootstrap replicates). The gold star marks the sensitivity-constrained operating point; the dashed line in \textbf{(b)} indicates chance-level precision.}\label{fig:roc_pr}
\end{figure}

Figure~\ref{fig:roc_pr} summarizes held-out test-set performance of the primary screening head at the fixed transferred threshold $\tau^{\star}=0.44$. Panel~(a) shows the ROC curve (AUROC = 0.933) and panel~(b) the precision--recall curve (AUPRC = 0.956); the gold star marks the operating point (sensitivity = 0.904, specificity = 0.773). Panel~(c) consolidates six operating-point metrics with 95\% confidence intervals from 2\,000 stratified bootstrap replicates.

The sensitivity of 0.904 exceeds the target constraint of 0.86 applied during threshold search, confirming that the screening objective of minimising missed bone-loss cases was achieved. The specificity of 0.773 implies that approximately 22.7\% of normal knees were referred for confirmatory DXA at this operating point.

\subsection{Ablation Study and Cross-Backbone Benchmark}\label{subsec:benchmarking}

\begin{figure}[t]
\centering
\includegraphics[width=\linewidth]{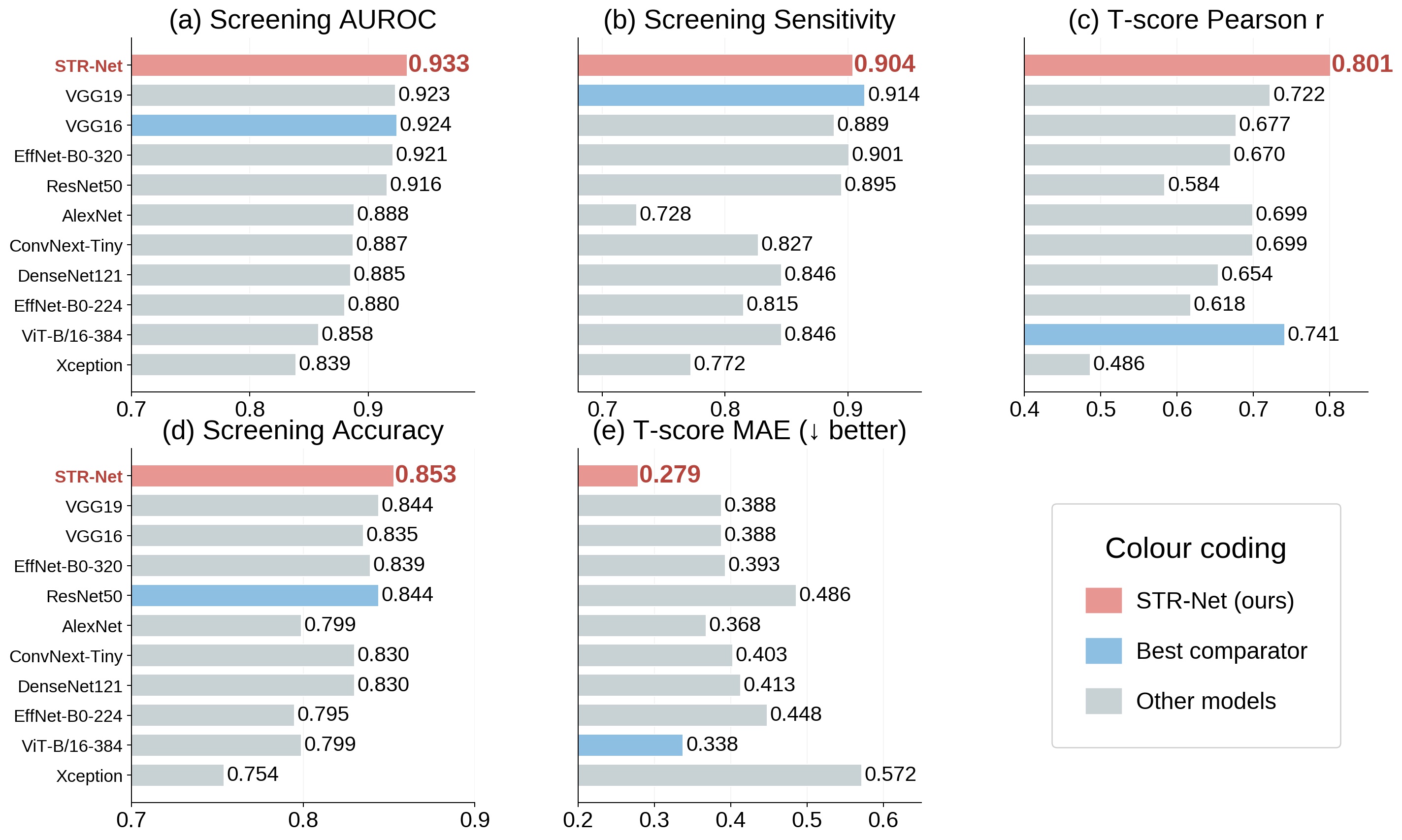}
\caption{Cross-backbone benchmark on the held-out test set (n=224); only the backbone feature extractor differs, while shared neck, TARR routing, and task heads remain fixed. \textbf{(a)}~Screening AUROC. \textbf{(b)}~Screening sensitivity. \textbf{(c)}~T-score Pearson correlation on the pilot T-score subset (n=31). \textbf{(d)}~Screening accuracy. \textbf{(e)}~T-score MAE (lower is better). Color coding follows the figure legend: STR-Net (ours), best comparator, and other models.} \label{fig:benchmark_summary}
\end{figure}

\begin{figure}[t]
\centering
\includegraphics[width=0.9\linewidth]{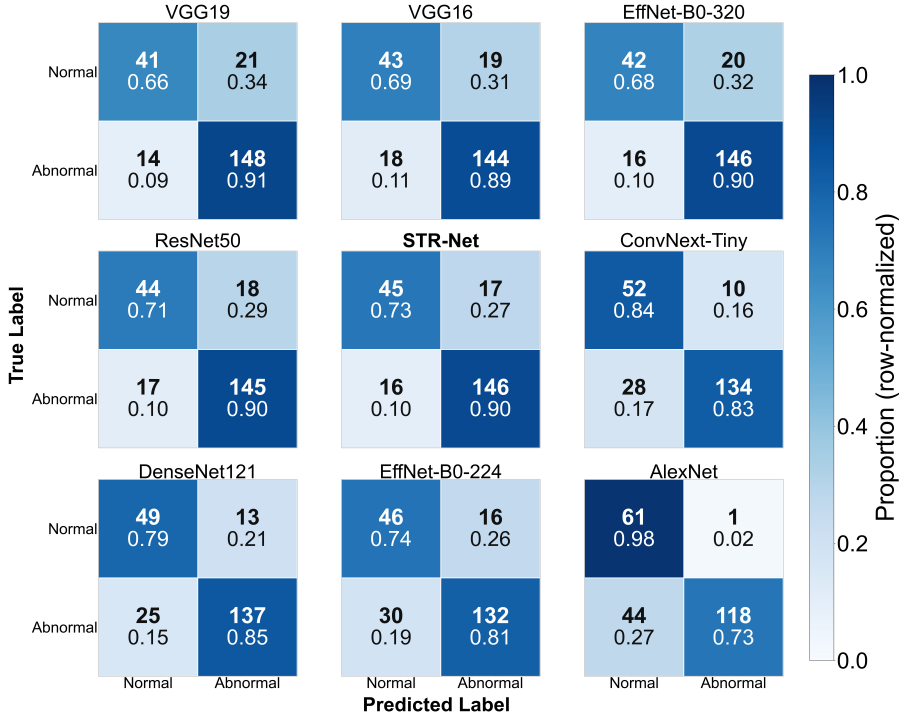}
\caption{Row-normalized confusion matrices for the compared backbones on the held-out test set. For each model, predicted probabilities are thresholded at the validation-selected sensitivity-constrained operating point (Section~\ref{subsec:evaluation}). Each cell reports count (top) and row proportion (bottom). In the STR-Net panel, counts are 45 true negatives, 17 false positives, 16 false negatives, and 146 true positives.}\label{fig:benchmark_confusion}
\end{figure}

All comparisons in this subsection use the same held-out test set (n=224). For each ablation stage and backbone, the operating point was tuned independently on validation data with the sensitivity-constrained rule in Section~\ref{subsec:evaluation}. We report (i) internal component ablation (Table~\ref{tab:ablation_progression}) and (ii) cross-backbone benchmarking with ten alternative feature extractors while keeping neck, TARR, and task heads fixed (Figs.~\ref{fig:benchmark_summary}--\ref{fig:benchmark_confusion}).

Table~\ref{tab:ablation_progression} shows monotonic B0$\rightarrow$B4 improvement in screening metrics. Severe-head AUROC appears from B2 and T-score RMSE from B3, as expected from output availability. Because stepwise gains are close to seed-level variance, results are interpreted as aggregate trend rather than strict component separation at this sample size.

\begin{table*}[b]
\caption{Ablation on the held-out test split ($n=224$ images): each row adds one component to the previous configuration. Values are mean $\pm$ standard deviation across 5 independent training runs with random seeds \{13, 42, 123, 456, 2024\}. AUPRC, AUROC, and screening sensitivity (Sens.) were computed on all held-out test images. Sev.\ AUROC is the ROC AUC of the severity head for Osteoporosis vs.\ Osteopenia (positive class: Osteoporosis) on the Bone-Loss subset ($n=136$). RMSE is the root mean squared error of the T-score branch on the held-out subset with available T-score labels ($n=31$; Section~\ref{subsec:tscore}). Dashes indicate outputs unavailable at that stage.}\label{tab:ablation_progression}
\centering
\small 
\setlength{\tabcolsep}{4pt} 
\begin{tabular}{@{} l r@{\,$\pm$\,}l r@{\,$\pm$\,}l r@{\,$\pm$\,}l r@{\,$\pm$\,}l r@{\,$\pm$\,}l @{}}
\toprule
\textbf{Stage} & 
\multicolumn{2}{c}{\textbf{AUPRC}} & 
\multicolumn{2}{c}{\textbf{AUROC}} & 
\multicolumn{2}{c}{\textbf{Sens.}} & 
\multicolumn{2}{c}{\textbf{Sev. AUROC}} & 
\multicolumn{2}{c}{\textbf{RMSE}} \\
\midrule
B0 baseline         & 0.930 & 0.010 & 0.916 & 0.013 & 0.895 & 0.017 & \multicolumn{2}{c}{--} & \multicolumn{2}{c}{--} \\
B1 + neck           & 0.937 & 0.009 & 0.920 & 0.012 & 0.897 & 0.016 & \multicolumn{2}{c}{--} & \multicolumn{2}{c}{--} \\
B2 + severe head    & 0.943 & 0.008 & 0.924 & 0.011 & 0.899 & 0.015 & 0.882 & 0.019 & \multicolumn{2}{c}{--} \\
B3 + T-score branch & 0.950 & 0.009 & 0.929 & 0.012 & 0.901 & 0.016 & 0.890 & 0.018 & 0.435 & 0.040 \\
B4 + TARR (STR-Net) & \textbf{0.956} & \textbf{0.008} & \textbf{0.933} & \textbf{0.012} & \textbf{0.904} & \textbf{0.015} & \textbf{0.898} & \textbf{0.018} & \textbf{0.418} & \textbf{0.036} \\
\bottomrule
\end{tabular}
\end{table*}

Figure~\ref{fig:benchmark_summary} reports the full cross-backbone comparison in the final figure: screening AUROC \textbf{(a)}, screening sensitivity \textbf{(b)}, T-score Pearson $r$ \textbf{(c)}, screening accuracy \textbf{(d)}, and T-score MAE \textbf{(e)}. STR-Net achieves the highest AUROC (0.933) and the lowest MAE (0.279); VGG19 has the highest screening sensitivity (0.914); and ViT-B/16-384 is the strongest comparator on Pearson $r$ (0.741).

Figure~\ref{fig:benchmark_confusion} complements curve-based metrics with thresholded confusion matrices. Under the sensitivity-constrained protocol, STR-Net shows a comparatively balanced recall/false-positive trade-off on this split.

\subsection{Severity Sub-Classification}\label{subsec:severity}

Within the Bone-Loss subset (n=136), the masked severity head (Osteopenia vs.\ Osteoporosis) achieved AUROC 0.898 and AUPRC 0.888 (Table~\ref{tab:severe_results}). At the selected operating point, sensitivity was 0.862, specificity 0.765, PPV 0.810, NPV 0.825, and MCC 0.629. These results are interpreted as within-Bone-Loss refinement rather than whole-cohort classification.

\begin{table}[h]
\normalsize
\begin{tabular*}{\linewidth}{@{}l@{\extracolsep{\fill}}cc@{}}
\toprule
\textbf{Metric} & \textbf{Value} & \textbf{95\,\% CI} \\
\midrule
\multicolumn{3}{@{}l}{\textit{Area-under-curve (threshold-free)}} \\[2pt]
\quad AUPRC                  & 0.888 & {[0.832, 0.944]} \\
\quad AUROC                  & 0.898 & {[0.844, 0.952]} \\[4pt]
\multicolumn{3}{@{}l}{\textit{Accuracy (at operating point)}} \\[2pt]
\quad Accuracy               & 0.818 & {[0.753, 0.883]} \\
\quad Balanced Accuracy      & 0.806 & {[0.732, 0.880]} \\[4pt]
\multicolumn{3}{@{}l}{\textit{Operating-point metrics}} \\[2pt]
\quad Sensitivity (Recall)   & 0.862 & {[0.783, 0.941]} \\
\quad Specificity            & 0.765 & {[0.659, 0.871]} \\
\quad PPV (Precision)        & 0.810 & {[0.724, 0.896]} \\
\quad NPV                    & 0.825 & {[0.726, 0.924]} \\
\quad F1-Score               & 0.837 & {[0.753, 0.921]} \\
\quad Youden's J             & 0.627 & {[0.496, 0.758]} \\[4pt]
\multicolumn{3}{@{}l}{\textit{Composite}} \\[2pt]
\quad MCC                    & 0.629 & {[0.486, 0.772]} \\
\bottomrule
\end{tabular*}
\caption{Held-out test-set performance of the severity head within the Bone-Loss subset ($n = 136$; 74 Osteoporosis, 62 Osteopenia). Osteoporosis is treated as the positive class. 95\,\% CIs estimated by stratified bootstrap ($B = 2\,000$). The severe head is the masked branch shown in Figs.~\ref{fig:architecture}--\ref{fig:tarr_detail}.}\label{tab:severe_results}
\end{table}

\subsection{Auxiliary T-Score Regression}\label{subsec:tscore}

On the test subset with T-score labels (n=31), the regression branch yielded Pearson $r=0.801$, MAE 0.279, and RMSE 0.347. This remains pilot-only evidence due to small sample size. Figure~\ref{fig:tscore_scatter} shows point-level agreement with the identity line and OLS fit; backbone-level Pearson and MAE comparisons are shown in Fig.~\ref{fig:benchmark_summary}(c,e).

\begin{figure}[h]
\centering
\includegraphics[width=0.9\linewidth]{Fig8_tscore_scatter.jpg}
\caption{Predicted versus reference T-score on the pilot test subset with available labels (n=31). Dashed line: identity ($y=x$); solid line: ordinary least-squares fit.}\label{fig:tscore_scatter}
\end{figure}

\subsection{Independent external validation}\label{subsec:external_results}

Table~\ref{tab:external_baseline} summarizes the external cohort (n=42, linked knee radiograph + DXA-based hip/spine BMD-derived T-score). The cohort was female-predominant (88.1\%), with mean age 63.1~$\pm$~15.2 years, BMI 26.2~$\pm$~4.2~kg/m$^2$, and Bone-Loss prevalence 71.4\% (30/42). Acquisition reflected routine clinical heterogeneity; scanner-level stratification was not feasible at this sample size.

\begin{table}[h]
\normalsize
\begin{tabular*}{\linewidth}{@{}l@{\extracolsep{\fill}}c@{}}
\toprule
\textbf{Characteristic} & \textbf{External cohort (n=42)} \\
\midrule
Age (years)                         & 63.1 $\pm$ 15.2 \\
Female sex, $n$~(\%)                 & 37 (88.1\%) \\
BMI (kg/m$^2$)                      & 26.2 $\pm$ 4.2 \\
Bone Loss (T-score $\leq -1.0$), $n$~(\%) & 30 (71.4\%) \\
\bottomrule
\end{tabular*}
\caption{Baseline characteristics of the independent external validation cohort with linked  DXA-based hip or spine BMD-derived T-scores. Continuous variables: mean~$\pm$~SD. Bone Loss denotes osteopenia or osteoporosis relative to young-adult norms (Section~\ref{subsec:dataset}).}\label{tab:external_baseline}
\end{table}

\begin{figure}[t]
\centering
\includegraphics[width=\linewidth]{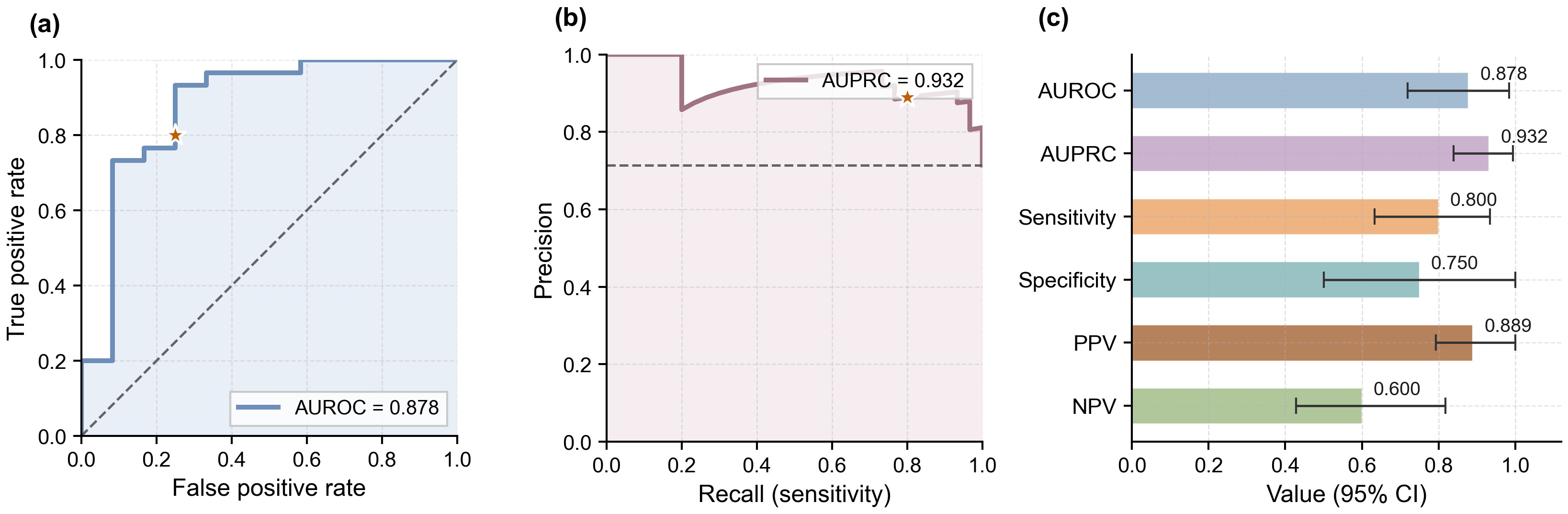}
\caption{Independent external cohort (n=42; 2{,}000 stratified bootstrap replicates). \textbf{(a)}~ROC curve for Bone Loss screening at the fixed development threshold $\tau^{\star}=0.44$ (gold star). \textbf{(b)}~Precision--recall curve; dashed line: cohort prevalence of Bone Loss. \textbf{(c)}~Point estimates with 95\,\% CI for AUROC, AUPRC, sensitivity, specificity, PPV, and NPV.}\label{fig:external_roc_pr}
\end{figure}

Figure~\ref{fig:external_roc_pr} reports external screening performance at fixed $\tau^{\star}=0.44$: AUROC 0.878 (95\,\% CI 0.719--0.983), AUPRC 0.932 (0.839--0.994), sensitivity 0.800 (0.633--0.933), specificity 0.750 (0.500--1.000), PPV 0.889 (0.793--1.000), NPV 0.600 (0.429--0.818), and accuracy 0.786. Compared with internal testing, discrimination and sensitivity were lower, consistent with domain shift.

\begin{figure}[t]
\centering
\includegraphics[width=\linewidth]{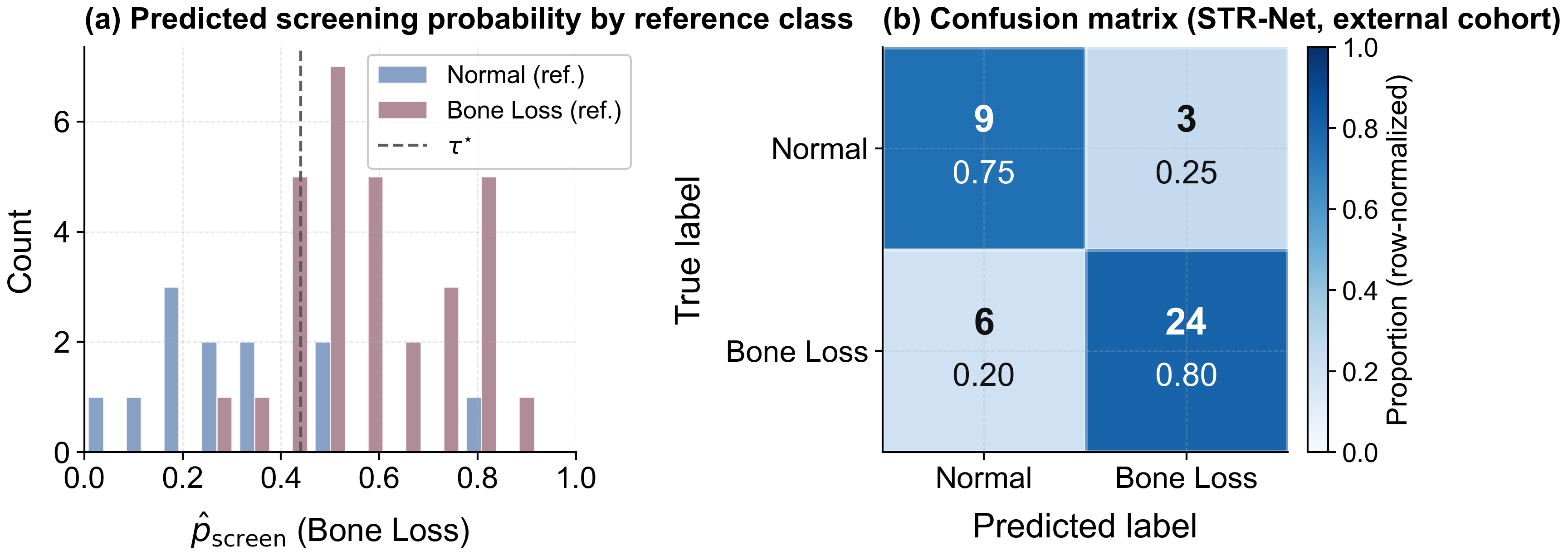}
\caption{External cohort analysis at $\tau^{\star}=0.44$. \textbf{(a)}~Histogram of predicted screening probabilities by DXA-derived reference class (dashed vertical line: $\tau^{\star}$). \textbf{(b)}~Confusion matrix for binary screening; cell shading maps the row-normalized proportion within each reference class, with counts (bold) and proportions shown in each cell.}\label{fig:external_calibration}
\end{figure}

Figure~\ref{fig:external_calibration} shows score distribution at $\tau^{\star}$ and confusion structure (24 TP, 6 FN, 3 FP, 9 TN; Brier score 0.178).

\begin{figure}[h]
\centering
\includegraphics[width=0.9\linewidth]{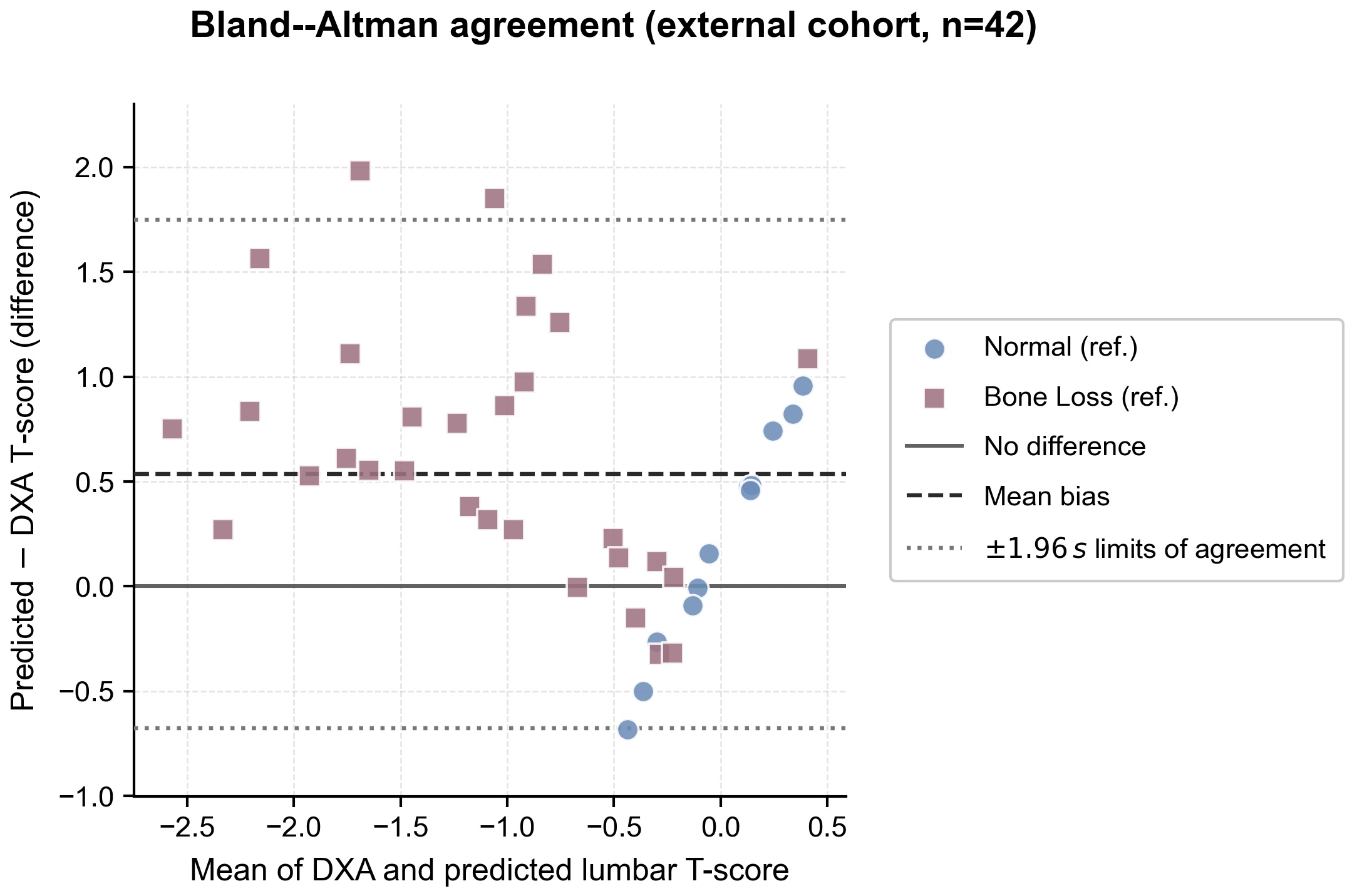}
\caption{Bland--Altman plot for the T-score branch on the external cohort ($n=42$): each point shows the mean of the predicted and DXA-reported hip/spine-derived T-score versus their difference (predicted minus DXA). Solid line: no difference; dashed: mean bias; dotted: approximate 95\,\% limits of agreement. Markers follow the screening reference category (Normal vs.\ Bone Loss).}\label{fig:external_tscore}
\end{figure}

Figure~\ref{fig:external_tscore} provides Bland--Altman agreement for T-score prediction versus DXA in the same cohort ($n=42$). Summary metrics were Pearson $r=0.766$, MAE 0.647, and RMSE 0.813.

\subsection{Lesion Localization Through Grad-CAM}\label{subsec:gradcam_results}

\begin{figure}[t]
\centering
\includegraphics[width=0.31\linewidth]{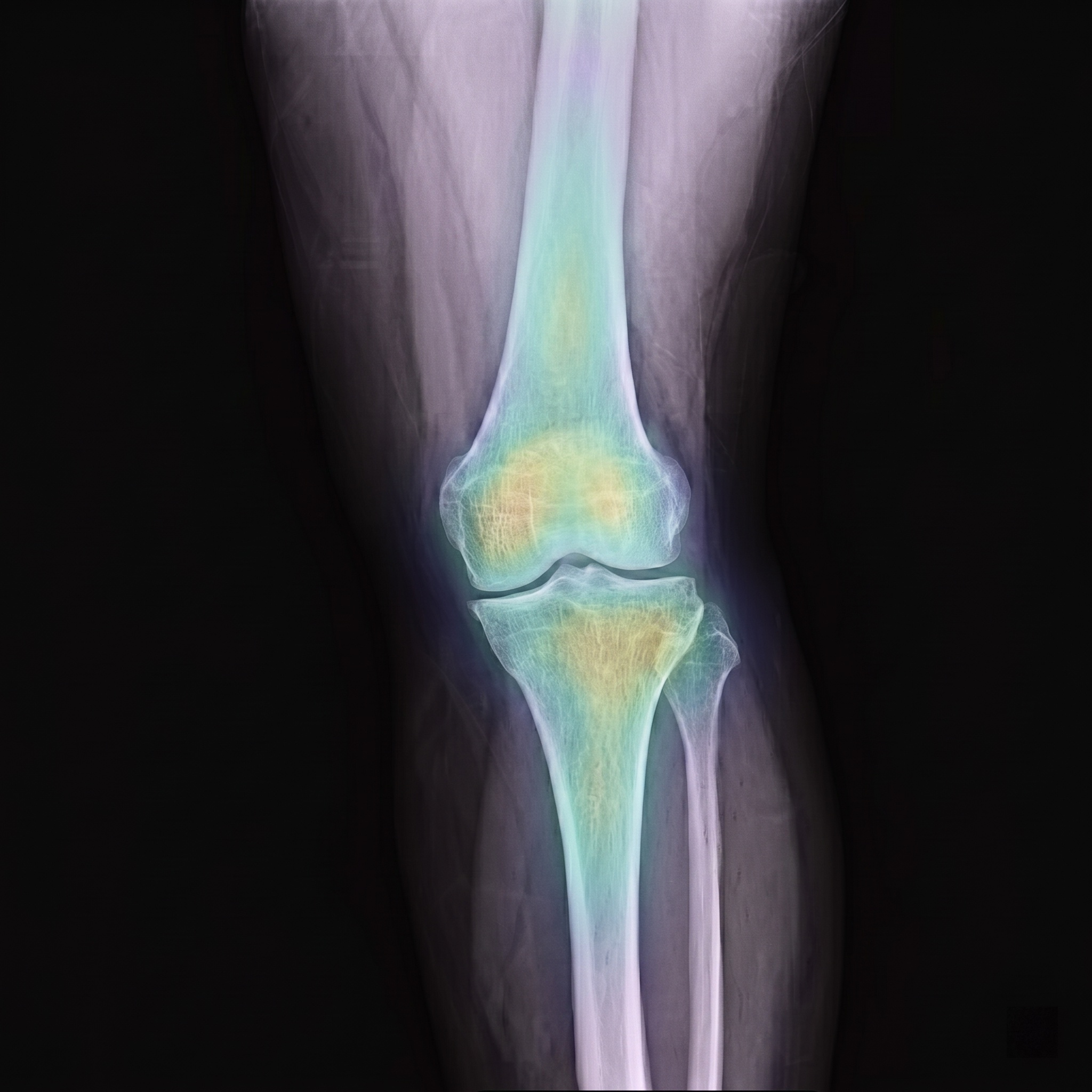}\hfill
\includegraphics[width=0.31\linewidth]{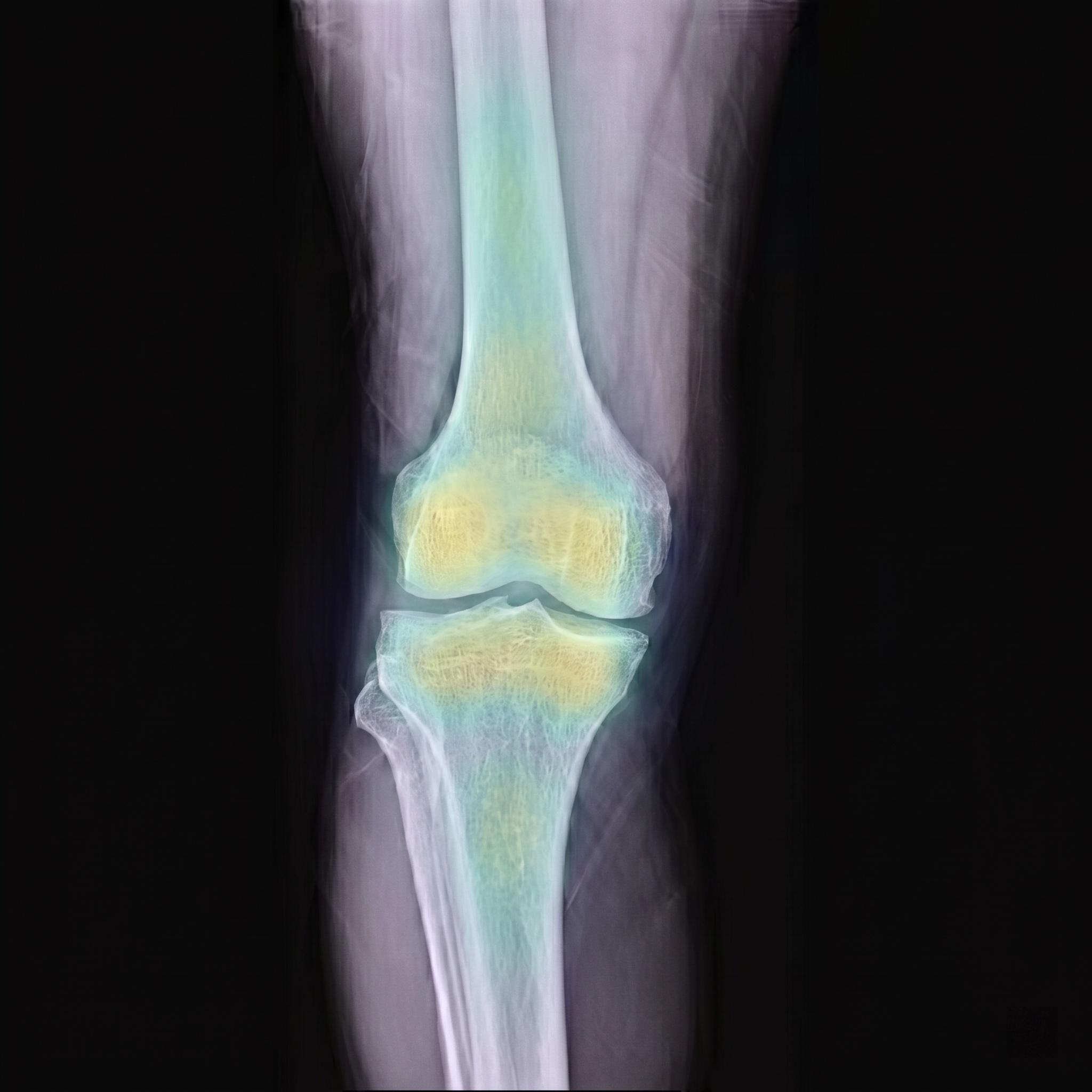}\hfill
\includegraphics[width=0.31\linewidth]{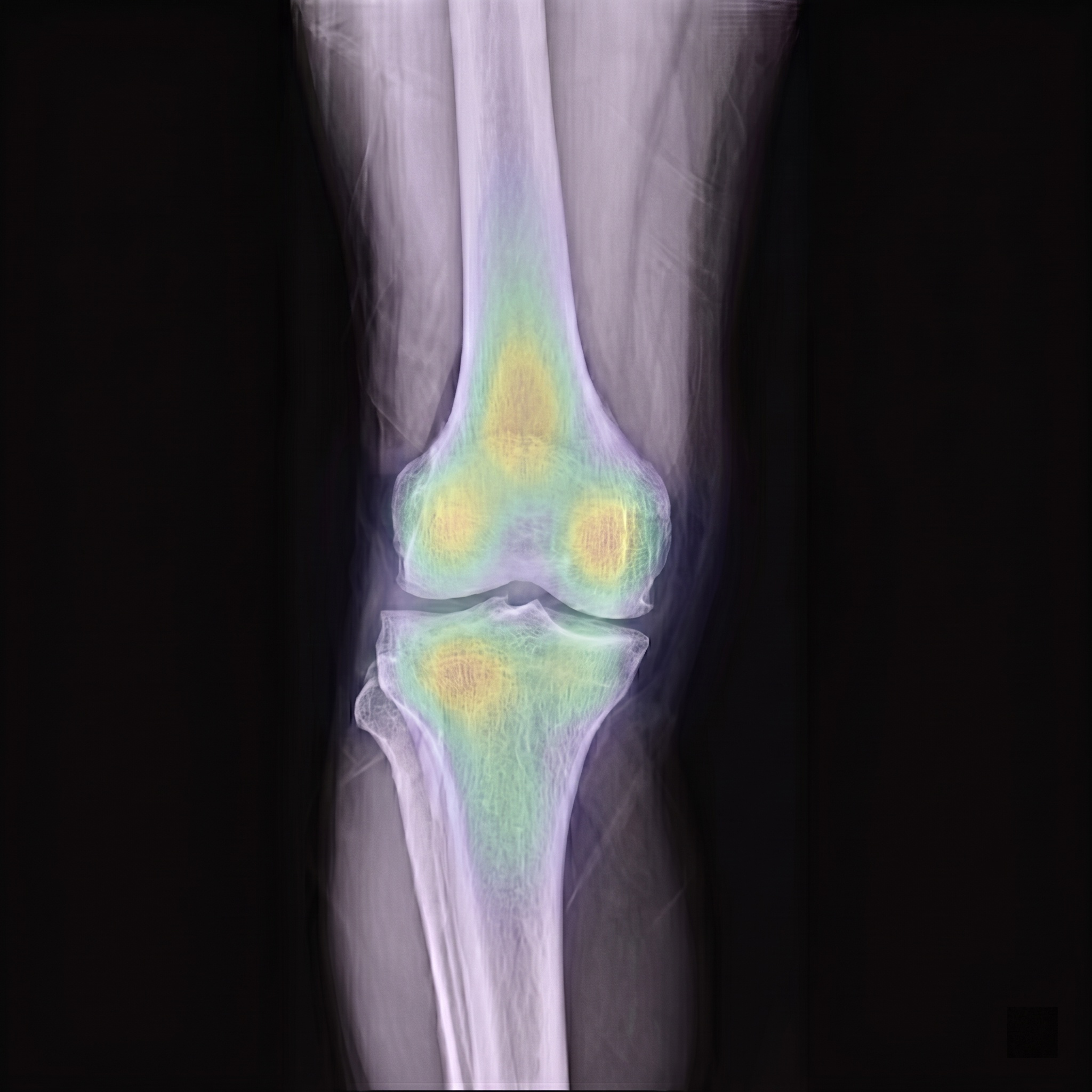}
\caption{Grad-CAM overlays for three representative test cases (left, center, right). Maps are computed from the last convolutional block of the ResNet-50 backbone; warmer colors indicate stronger activation.}\label{fig:gradcam}
\end{figure}

Representative Grad-CAM overlays are shown in Fig.~\ref{fig:gradcam}. Across the three example cases, activation is concentrated in periarticular cancellous regions of the distal femur and proximal tibia, with limited edge/soft-tissue emphasis. This pattern is anatomically plausible but remains qualitative.

\section{Discussion}\label{sec:discussion}

\subsection{Main Findings}\label{subsec:main_findings}

This study presents a multi-task pipeline for opportunistic bone-loss screening from routine knee radiographs. Internal test performance was strong (AUPRC 0.956, AUROC 0.933; sensitivity 0.904, specificity 0.773 at $\tau^{\star}=0.44$). Severity sub-classification reached AUROC 0.898. T-score regression remained exploratory (n=31, Pearson $r=0.801$). External DXA-based validation (n=42) showed AUROC 0.878, sensitivity 0.800, specificity 0.750, and T-score correlation $r=0.766$ under the same threshold.

\subsection{Clinical workflow, threshold policy, and deployment}\label{subsec:clinical_translation}

The practical value is workflow integration: the model uses radiographs already acquired in routine care, potentially flagging high-risk patients for DXA without added imaging burden. This aligns with opportunistic imaging-informatics use cases~\cite{pickhardt2013,yamamoto2020} and may be particularly relevant in high-volume knee-radiograph settings~\cite{bedson2007kneexray}.

Sensitivity-constrained thresholding was chosen for screening, where missed positives are usually costlier than false referrals~\cite{shepstone2018}. On internal testing, this policy achieved sensitivity 0.904 and specificity 0.773 at fixed $\tau^{\star}=0.44$. Resulting PPV and referral workload are prevalence-dependent and should not be extrapolated directly to lower-prevalence community settings.

Figure~\ref{fig:benchmark_confusion} illustrates this trade-off: higher recall typically increases false positives, whereas stricter thresholds reduce referrals but miss more cases. Final thresholding should be calibrated to local prevalence and health-economic priorities.

Clinical deployment would still require substantial engineering (DICOM ingestion, latency validation, reporting integration) and formal regulatory pathways. The present study establishes technical feasibility, not deployment readiness.

\subsection{Auxiliary outputs, robustness, and interpretability}\label{subsec:auxiliary_trust}

Auxiliary outputs may support triage (severity AUROC 0.898), but T-score prediction remains pilot-level. Ablation trends support the multi-task design directionally, yet component-level effects are modest relative to seed variance. Preprocessing choices were intended to reduce reliance on acquisition artefacts. Grad-CAM maps appeared anatomically plausible (periarticular cancellous emphasis), but this evidence is qualitative and cannot replace quantitative localization validation~\cite{topol2019,litjens2017,riggs1981jci}.

\subsection{Limitations}\label{subsec:limitations}

This study has several limitations that should be acknowledged:

\begin{enumerate}
  
  \item \textbf{Reference-standard heterogeneity in the development cohort.} Development labels are source-repository-assigned and not uniformly DXA-based. Stage~1 is linked to calcaneal QUS and directory-level class assignment, while Stage~2 is a redistribution that includes Stage-1-derived material. As a result, held-out development metrics quantify consistency with repository labels rather than calibration against a single uniform DXA reference standard.

  \item \textbf{Residual identity-leakage uncertainty in Stage~1.} Stage~1 lacks verifiable source-side patient identifiers, so image-level splitting cannot exclude the possibility that multiple radiographs from the same patient appear across partitions. This uncertainty remains first-order when interpreting internal performance.

  \item \textbf{Ablation margins relative to seed variance.} Component-wise margins in the ablation study are modest relative to seed-level variance; clear separation of shared neck, severe head, T-score branch, and TARR contributions will require larger held-out resamples or multi-site replication. 

  \item \textbf{Limited T-score evaluation.} Only 31 test samples carried T-score labels, which is insufficient for robust statistical conclusions about T-score prediction; the regression branch is therefore presented as pilot feasibility evidence.

  \item \textbf{Qualitative-only interpretability.} The Grad-CAM analysis (Fig.~\ref{fig:gradcam}) is restricted to visual inspection of representative cases. Quantitative evaluation of localization fidelity (for example overlap with independently defined regions of interest) has not been performed.

  \item \textbf{No prospective validation.} The study is retrospective; prospective evaluation of effects on DXA referral rates and time-to-diagnosis is still required.
\end{enumerate}

\section{Conclusions}\label{sec:conclusions}

STR-Net achieved strong internal screening performance on routine knee radiographs (AUPRC 0.956; AUROC 0.933; sensitivity 0.904; specificity 0.773 at $\tau^{\star}=0.44$), with supportive severity stratification and exploratory T-score outputs. External DXA-based validation (n=42) showed lower but retained discrimination (AUROC 0.878), consistent with partial transfer under dataset shift. Overall, results support feasibility but not deployment-grade validation.

Key limitations are heterogeneous development labels, limited external sample size, small T-score subset, and qualitative-only interpretability analysis. Future work should prioritize larger multi-site validation, quantitative localization assessment, and prospective workflow-impact studies.

\backmatter

\bmhead{Acknowledgements}

Not applicable.

\section*{Declarations}

\begin{itemize}
  \item \textbf{Funding:} This work was supported by the National Natural Science Foundation of China (NSFC) Youth Program (Category C; Grant No. 82402389), the Natural Science Foundation of Liaoning Province General Program (Grant No. 2025-MS-027), and the Fundamental Research Funds for the Central Universities (Grant No. N25LPY051).
  \item \textbf{Conflict of interest:} The authors declare that they have no conflict of interest.
  \item \textbf{Ethics approval and consent to participate:} Analysis of fully de-identified, publicly redistributed imaging data from the two Kaggle repositories was conducted under the authors' institutional research ethics framework governing secondary analysis of public data, which does not require separate IRB submission for re-analysis of such material; the original data collection and ethical approvals are documented in the respective source repositories and were accessed in accordance with their stated terms of use. For the independent external validation cohort (retrospective linked knee radiographs and DXA-based hip and/or spine BMD/T-score reports at a collaborating hospital), ethical approval was granted by the institutional Research Ethics Committee (Approval No.: 2025PS1425K), and the study was conducted in accordance with the Declaration of Helsinki. 
\item \textbf{Consent for publication:} Not applicable.
 \item \textbf{Data and code availability:} Analysis code (preprocessing, training, and evaluation utilities) is publicly available at \url{https://github.com/TadejPogacar105/Osteo_Screening}. The image datasets used in this study are publicly available as follows: Stage~1 (``Osteoporosis X-ray'') is available via Kaggle at \url{https://www.kaggle.com/datasets/majesticd/osteoporosis-x-ray}, with the underlying public dataset archived as \emph{Knee X-ray Osteoporosis Database} (Mendeley Data; DOI: \url{https://doi.org/10.17632/fxjm8fb6mw.2}). Stage~2 (``Multi-class Knee Osteoporosis X-ray Dataset'') is available via Kaggle at \url{https://www.kaggle.com/datasets/mohamedgobara/multi-class-knee-osteoporosis-x-ray-dataset}; an associated publication describing this dataset is Sarhan \emph{et al.}, \emph{Knee Osteoporosis Diagnosis Based on Deep Learning} (DOI: \url{https://doi.org/10.1007/s44196-024-00615-4}). Train/validation/test split files (Stage~1 image-level; Stage~2 patient-level) and trained model weights will be added to the same GitHub repository where not already present; until then, these materials will be provided upon reasonable request for reproducibility verification.

  \item \textbf{Use of generative AI and large language models (LLMs):} The authors did not use generative AI or LLMs in the preparation of this manuscript.
  \item \textbf{Author contribution:} Following the CRediT taxonomy: Zhaochen Li (conceptualization, methodology, software, formal analysis, investigation, data curation, visualization, writing---original draft); Xinghao Yan (methodology, software, formal analysis, investigation, validation, visualization, writing---original draft); Runni Zhou (investigation, validation, data curation, writing---review and editing); Xiaoyang Li (investigation, resources, data curation, writing---review and editing); Chenjie Zhu (validation, formal analysis, visualization, writing---review and editing); Gege Wang (data curation, data cleaning, investigation); Yu Shi (resources, data curation, clinical data acquisition, medical supervision); Lixin Zhang (resources, data curation, clinical data acquisition, medical supervision); Rongrong Fu (validation, investigation, writing---review and editing); Liehao Yan (supervision, project administration, writing---review and editing); Yuan Chai (conceptualization, supervision, project administration, writing---review and editing, funding acquisition, correspondence). Zhaochen Li and Xinghao Yan contributed equally as co-first authors. Yuan Chai is the corresponding author. All authors read and approved the final manuscript.
\end{itemize}

\IfFileExists{main_blinded.bbl}{\input{main_blinded.bbl}}{\bibliography{references}}
\end{document}